%% file: main.tex
\documentclass{article}

\usepackage[preprint]{neurips_2026}

\usepackage[utf8]{inputenc} % allow utf-8 input
\usepackage[T1]{fontenc}    % use 8-bit T1 fonts
\usepackage{hyperref}       % hyperlinks
\usepackage{url}            % simple URL typesetting
\usepackage{booktabs}       % professional-quality tables
\usepackage{amsfonts}       % blackboard math symbols
\usepackage{nicefrac}       % compact symbols for 1/2, etc.
\usepackage{microtype}      % microtypography
\usepackage{xcolor}         % colors
\usepackage{amsmath}
\usepackage{graphicx}
\usepackage{wrapfig}
\usepackage{adjustbox}
\usepackage{algorithm}
\usepackage{algpseudocode}
\usepackage{booktabs}
\usepackage{multirow}
\usepackage{enumitem} 
\usepackage{array}
\usepackage{caption}
\usepackage{hyperref}
\usepackage{makecell}
\usepackage{tcolorbox}
\usepackage{subcaption}
\usepackage{listings}
\usepackage{graphicx}
\usepackage{wrapfig,lipsum,booktabs}
\usepackage[table]{xcolor}
\usepackage{algpseudocode}
\usepackage{xspace}
\usepackage{amsmath,amssymb}
\usepackage{bbm}

\usepackage{pifont} % 在导言区加载包
\newcommand{\method}{\textsc{EvalBuilder}\xspace}

\title{\method: Can LLMs Build Efficient Benchmarks?}

\author{%
Huaisheng Zhu$^{1}$\thanks{Equal contribution.},
\textbf{Hangfan Zhang}$^{2}$\footnotemark[1], 
Teng Xiao$^{4,5}$, 
\textbf{Siyuan Xu}$^{2}$,
Shijie Zhou$^{3}$, \\
\textbf{Zhimeng Guo}$^{2}$,
\textbf{Zhengyu Chen}$^{1}$\thanks{Corresponding authors.},
\textbf{Jingang Wang}$^{1}$\footnotemark[2] \\
$^1$ Meituan \quad
$^2$ Pennsylvania State University \quad
$^3$ University at Buffalo \\
$^4$ Allen Institute for AI (AI2) \quad
$^5$ University of Washington \\
\texttt{zhuhuaisheng@meituan.com}
}

\begin{document}

\maketitle

\begin{abstract}
  Large language model (LLM) agents are increasingly used for long-horizon research and engineering, yet evaluating these capabilities remains expensive, time-consuming, and heavily dependent on human expertise. This motivates a natural question: can LLM agents themselves construct efficient benchmarks that preserve the conclusions of costly reference evaluations? We study this capability by formulating automatic benchmark development as a meta-evaluation problem, where an agent is given a task and an expensive reference benchmark and requested to construct a lower-cost proxy benchmark that faithfully distinguishes among candidate models and solutions. Based on this formulation, we introduce \method, a framework for evaluating such benchmark-generation capabilities along three key dimensions: fidelity, efficiency, and robustness. Specifically, we assess whether generated benchmarks preserve model rankings, generalize to held-out models and freshly collected responses, and provide well-calibrated and discriminative scores while substantially reducing evaluation cost. We further provide an iterative environment in which agents can refine their benchmarks under a fixed time budget. Our framework provides a systematic testbed and benchmark for understanding when current LLM agents can develop useful evaluators and how benchmark construction itself can be increasingly automated. Our code is available
  at \href{https://github.com/huaishengzhu/EvalBuilder}{Github}.
\end{abstract}

\input{sections/introduction-v2}

\input{sections/related-works}

\input{sections/prelim}

\input{sections/method-v2}

\input{sections/experiment}

\input{sections/conclusion}
\bibliography{ref}
\bibliographystyle{plain}

%%%%%%%%%%%%%%%%%%%%%%%%%%%%%%%%%%%%%%%%%%%%%%%%%%%%%%%%%%%%

\input{sections/appendix}

\end{document}

%% file: sections/introduction-v2.tex
\section{Introduction}

Large language models (LLMs) are increasingly deployed as agents for long-horizon research and engineering workflows, where they iteratively propose ideas, implement solutions, evaluate outcomes, and refine their approaches~\cite{yao2022react,singh2025openai, xu2026deepseek,team2026kimi,yang2025qwen3}. Despite this progress, current LLMs still exhibit some limitations in autonomous discovery and engineering execution, particularly in complex long-horizon settings~\cite{yang2026programbench,fei2026agents,li2026beyond}. These limitations motivate the development of stronger models with more capable agentic behaviors. A key prerequisite for such progress is rigorous evaluation: carefully designed benchmarks are needed to assess long-horizon research and engineering capabilities, diagnose model weaknesses, and provide informative signals for further improvement. Accordingly, a growing number of benchmarks have been introduced to evaluate these capabilities and support training of more capable LLM agents~\cite{jimenez2023swe,merrill2026terminal,lyu2026mls,barres2025tau,chan2025mle,rank2026posttrainbench}.

However, existing benchmarks for agent tasks are often cumbersome and costly to run, as evaluating long-horizon capabilities typically requires agents to operate for extended periods and consume substantial computational resources. For example, evaluating models on complex research and long-horizon engineering tasks may require half a day or even a full day of agent execution per task, while a single evaluation run can incur costs on the order of tens of thousands of dollars~\cite{zhu2026edgebench,li2026beyond}. Also, autonomous agents assume a larger role in research and engineering workflows, this reliance on manually designed evaluation increasingly becomes a major bottleneck to scaling model development. More importantly, constructing reliable evaluations for long-horizon tasks is itself costly and labor-intensive. High-fidelity evaluation often requires specialized sandbox environments, comprehensive automated test suites, domain-specific verification procedures, and substantial computational resources. Beyond infrastructure, human experts must determine which behaviors should be measured, design evaluation cases that distinguish genuine capability improvements from superficial shortcuts, and continually revise the protocol as new failure modes emerge. With the recent advances in long-horizon reasoning and the growing use of LLMs as assistants for complex tasks, an important research question naturally arises: \textit{to what extent can LLMs develop efficient and effective benchmarks that reduce human effort, and how should such benchmark-generation capabilities be evaluated?}

% This question leads to a meta-evaluation problem that is largely absent from conventional agent benchmarks. An object-level benchmark measures whether an agent can solve a task. A meta-level benchmark must instead measure whether the agent can develop the instrument by which future solutions to that task will be judged. 
To answer this question, we introduce \method, a framework for evaluating the ability of LLM agents to develop efficient and useful benchmarks. However, the main challenge in building such an evaluation pipeline lies in assessing generated benchmark quality from LLM agents, which often requires substantial human expertise and domain-specific judgment. To address this, we first formulate a setting for evaluating the ability of LLMs to construct benchmarks. In this setting, given a task and access to a cumbersome and expensive reference evaluation benchmark, an agent is required to construct a lower-cost benchmark whose judgments remain well aligned with the reference benchmark across a diverse set of candidate solutions. We further propose a multidimensional evaluation framework that assesses generated benchmarks from several complementary perspectives. Sepcifically, a poorly generated benchmarks may misrank candidate solutions, overfit to a narrow set of visible examples, or reward superficial shortcuts that fail to generalize to the intended task. Therefore, automatically generated benchmarks should be evaluated not only for their efficiency, but also for how faithfully they reproduce the judgments of the reference evaluation, how consistently they rank diverse candidate solutions, and how reliably they generalize beyond the examples observed during construction. To meet these criteria, we assess ranking fidelity, generalization to held-out models, and score calibration and resolution. Moreover, we provide an environment where agents iteratively refine tasks and verifiers under a fixed time budget, select their final benchmark from saved iterations, and submit it for independent evaluation using freshly collected model responses.

In summary, our work makes the following contributions toward building an agent framework for constructing efficient and useful benchmarks that reduce evaluation cost:
\begin{itemize}[leftmargin=*, noitemsep, topsep=0.4pt]
\item We formulate automatic benchmark development as a distinct capability of research agents, where agents are tasked with constructing efficient proxy benchmarks from an expensive target benchmark while preserving its ability to reliably distinguish the performance of different models.
\item We introduce \method, an evaluation pipeline designed to efficiently assess the ability of LLMs to construct effective and computationally efficient benchmarks.
\item We organize our assessment around fidelity, efficiency, and robustness, providing a unified framework for understanding when current LLM agents can construct useful evaluators, how their generated benchmarks fail, and how these benchmarks differentiate among frontier models.
\end{itemize}

% The central question is whether LLM agents can construct evaluators that deliver useful and trustworthy feedback for subsequent search and optimization; task-solving performance alone is insufficient to answer this question. 

% A natural next step is to ask whether an agent can construct a lightweight benchmark or evaluator for the task it is trying to solve. 

% However, a benchmark that is merely inexpensive or superficially plausible is insufficient. 

% A poorly designed evaluator may misrank candidate solutions, overfit to a limited set of visible examples, or reward shortcuts that fail to generalize to the intended task. Accordingly, automatically generated benchmarks must be assessed not only by their efficiency, but also by how faithfully they reproduce the decisions of the reference evaluation and how reliably they generalize beyond the examples used during construction.

% This formulation separates benchmark development from task solving, enabling systematic comparison of agents along three complementary dimensions: (1) fidelity to the reference evaluation, (2) evaluation efficiency, and (3) robustness to candidate solutions that were unavailable during benchmark construction.

%% file: sections/related-works.tex
\section{Related Works}

\label{sec:related}

\paragraph{Agents that conduct research and engineering.} LLMs have recently demonstrated strong capabilities on long-horizon tasks, particularly in complex research and engineering workflows~\cite{jimenez2024swe, openai2024swebenchverified,yang2026programbench,fei2026agents,li2026beyond}. Specifically, SWE-bench evaluates whether language models can resolve real GitHub issues in executable repositories \citep{jimenez2024swe}; SWE-bench Verified is a human-screened subset intended to remove underspecified or incorrectly tested instances \citep{openai2024swebenchverified}. Terminal-Bench extends agent evaluation to realistic command-line tasks \citep{merrill2026terminal,terminalbench2026v21}. Going beyond engineering agents, LLMs are increasingly being developed as research agents capable of conducting research and tackling complex scientific problems. For example, MLAgentBench evaluates iterative machine-learning experimentation \citep{huang2023mlagentbench}, PaperBench evaluates replication of AI research \citep{starace2025paperbench}, and the AI Scientist studies open-ended automation of the research process \citep{lu2024ai}. To measure these complex capabilities, it is essential to develop benchmarks that can reliably evaluate model performance and provide informative feedback for further improving model capabilities. However, constructing such benchmarks often requires substantial human effort and domain expertise, motivating the use of LLMs to automate or assist the benchmark development process. While prior work has extensively explored the use of LLMs to accelerate or automate various stages of model development—including data construction and filtering~\cite{kulikov2026autodata}, pretraining architecture design~\cite{karpathy2026autoresearch}, post-training~\cite{rank2026posttrainbench}, and agent harness design~\cite{lee2026meta,zhang2026self,lin2026agentic,wang2026rethinking,chen2026co}—the use of LLMs for automated benchmark development remains relatively underexplored. Therefore, in this work, we study the ability of LLMs to automatically design benchmarks, with the goal of reducing human effort and accelerating future model development.

% Dynabench couples model failures with human-in-the-loop data collection \citep{kiela2021dynabench}. More recent systems use coordinated agents to plan, generate, verify, and evaluate benchmark items \citep{butt2025benchagents}, generate code-task benchmarks alongside LLM judges \citep{farchi2024automatic}, or evolve test cases from multi-model feedback \citep{liu2025arenabencher}. These systems primarily ask whether automatic generation yields useful benchmark content. These efforts evaluate an agent under an evaluator supplied by benchmark designers. \method instead makes construction of the evaluator itself the object of study. 

\paragraph{Efficient Benchmark Buildup.} LLMs are used to resolve complex tasks which also make their evaluation more expansive. Accordingly, a growing body of work has explored using lightweight proxy benchmarks to predict performance on other benchmarks~\cite{spangher2025chatbot,polo2402tinybenchmarks,zhou2026lost,kipnis2407metabench,zhang2026sparseeval}. All of these methods operate within a single benchmark distribution and overlook the challenge of reliably predicting performance on an agentic benchmark using an external pool of heterogeneous, low-cost evaluation benchmarks. To address this limitation, several recent approaches leverage lightweight benchmarks from diverse domains to estimate performance on complex agentic tasks with highly heterogeneous distributions~\cite{song2026pace,zeng2026you,zhang2026benchmark}. However, these approaches primarily rely on conventional machine learning methods to predict performance using existing lightweight datasets from other domains, which may generalize poorly to new benchmarks that target capabilities not well represented by prior evaluations. Moreover, existing benchmark datasets may provide only limited coverage of the diverse and rapidly evolving capabilities exhibited by modern LLMs. To address these limitations, we propose leveraging LLMs to generate new evaluation questions, rather than selecting from existing benchmarks, and to analyze agent trajectories, enabling the construction of benchmarks that better capture the complexity and diversity of agentic task distributions.

 % \method asks a complementary meta-evaluation question: given a fixed reference ranking, can a single design agent produce both novel items and an executable verifier whose scores preserve that ranking, including for candidate models hidden during construction?

% \paragraph{Evaluator reliability.}
% LLM-as-a-judge work demonstrates the utility of model-based comparison while documenting position, verbosity, and self-preference biases \citep{zheng2023judging}; LLMBar supplies adversarial examples for instruction-following evaluators \citep{zeng2024llmbar}. These findings motivate evaluating the verifier together with its tasks instead of assuming that a plausible rubric is faithful. Our held-out-candidate and confirmation stages test two different forms of reliability: transfer beyond the candidates used for design, and persistence when responses are recollected independently.

%% file: sections/prelim.tex
\section{Preliminaries}
\label{sec:formulation}

We consider a set of candidate models $\mathcal{M}$ evaluated on an expensive reference benchmark, with reference score $r_m\in[0,1]$ for each model $m$. The benchmark-construction task is to develop a compact evaluation that reproduces their relative performance on this reference.

The agent constructs a benchmark $\mathcal{B}=(\mathcal{T},v)$, where $\mathcal{T}=\{t_i\}_{i=1}^{N}$ is a set of newly designed tasks and $v$ is an executable verifier. Each task contains a prompt and, optionally, an agent-authored reference answer or scoring rubric. Both the task set and the verifier are editable: the agent can add, remove, or revise tasks and modify their scoring logic throughout construction. The supplied reference results serve as fixed targets for assessing these changes. Given model $m$'s response $y_{m,i}$ to task $t_i$, the verifier returns a score $v(t_i,y_{m,i})\in[0,1]$. The model's benchmark score is:
\begin{equation}
    \hat r_m = \frac{1}{N}\sum_{i=1}^{N} v(t_i,y_{m,i}).
    \label{eq:benchmark-score}
\end{equation}
The primary objective is to preserve the ordering induced by the reference scores $\{r_m\}$, with score calibration and separation assessed as additional properties. We partition $\mathcal{M}$ into disjoint visible and held-out sets, $\mathcal{M}_{\mathrm{vis}}$ and $\mathcal{M}_{\mathrm{held}}$. During construction, the agent receives reference scores and trajectories for visible models and can query these models to jointly refine $\mathcal{T}$ and $v$ under a shared wall-clock budget. Held-out models are excluded from the construction inputs and are evaluated after the benchmark is submitted. Final assessment uses freshly collected responses from both groups to measure reference agreement and generalization to new candidates.

%% file: sections/method-v2.tex
\section{Method}
\label{sec:method}

\begin{figure}[t]
    \centering
    \includegraphics[width=\linewidth]
        {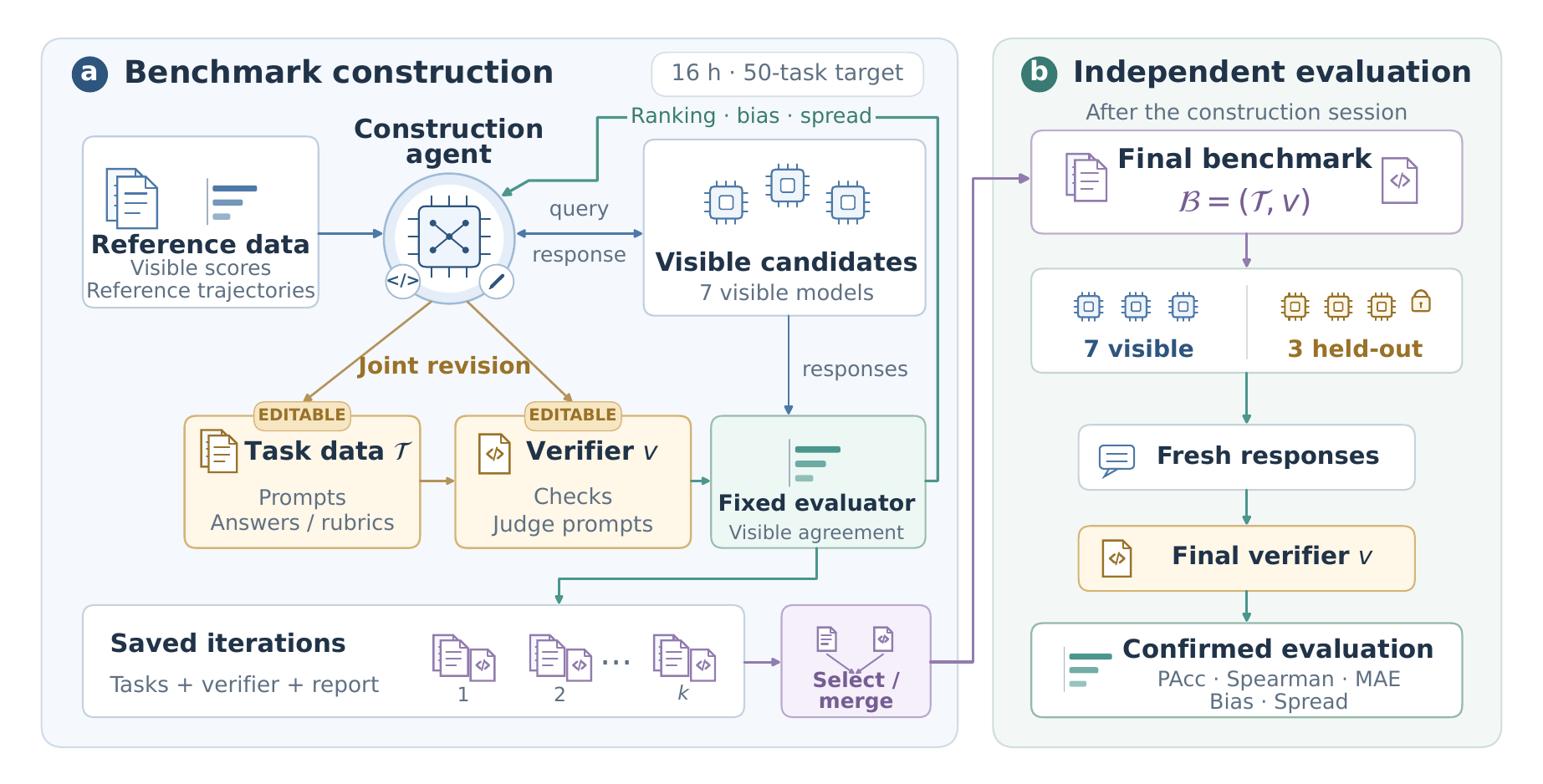}
    \caption{Overview of \method for building efficient and effective Benchmarks.}
    \label{fig:method}
\end{figure}

\method asks an agent to construct a compact benchmark that preserves the relative performance of models on an expensive reference benchmark. The agent designs a set of tasks together with an executable verifier, using reference results from visible models to guide its choices. We evaluate the resulting benchmark by its agreement with the reference evaluation, including on held-out models that were unavailable during construction. The overview of \method is shown in Figure~\ref{fig:method}.

\subsection{Agent Workspace and Benchmark Construction}
\label{sec:method:construction}

Each run provides the construction agent with a persistent workspace, tools for reading files and executing code, and an API for querying visible candidate models. The workspace contains an editable benchmark, supplied reference inputs, and a history of completed iterations.

\paragraph{Editable benchmark.}
The agent directly controls two artifacts: the task data (\texttt{responses.json}) and the verifier (\texttt{verify.py}). Each task record contains a prompt, an optional reference answer or rubric, and the collected model responses. The agent can add, remove, or rewrite tasks, revise their reference answers and rubrics, and refresh response records through model calls. It can also rewrite the verifier, including programmatic checks, judge prompts, partial-credit rules, and score calibration, while retaining the required per-response interface and $[0,1]$ output range. Task data and scoring code are therefore optimized jointly throughout construction. 

\paragraph{Reference inputs and working files.}
Visible models' aggregate reference scores (\texttt{truth.json}), reference trajectories, and the supplied evaluator (\texttt{evaluate.py}) provide the common basis for construction and feedback. The agent is instructed to leave these inputs unchanged. The trajectories include original task prompts, model actions, and success or failure outcomes, which the agent can analyze to identify recurring challenges and differences between models. Draft tasks and helper scripts for data generation, response collection, or scoring are kept in a separate scratch directory (\texttt{tmp\_operations/}). The current task data and verifier in the workspace root define the working benchmark.

\paragraph{Construction workflow.}
The initial round asks the agent to turn its analysis of the reference material into a working benchmark. It writes new tasks with variation in task type, domain, and difficulty, and specifies how their responses should be graded. For example, coding tasks may target debugging, implementation, or reasoning about edge cases. Task design and verifier design are coupled: the agent specifies both the behavior a task elicits and the evidence used to judge its quality.

The agent queries each visible model on the generated prompts and scores the collected responses using exact matching, programmatic checks, model-based judging, or a combination of these strategies. Missing or empty responses are flagged for retry, so collection failures can be addressed before interpreting differences in model performance. The agent then runs the supplied evaluator to obtain feedback for the next revision.

\subsection{Budgeted Refinement and Final Submission}
\label{sec:method:iteration}

Successive rounds continue the same agent session, allowing the agent to build on its previous analysis and saved benchmarks. A fixed evaluator aggregates the verifier's scores according to Equation~\eqref{eq:benchmark-score} and compares $\hat r_m$ with the reference score $r_m$ across visible models. It returns feedback on ranking agreement, score bias, and score spread. Each revision begins with a diagnosis of the current benchmark. The agent reviews incorrectly ordered model pairs, tasks that give all models similar scores, and responses whose grades appear unreliable. It then makes targeted changes to the tasks or verifier and reruns the evaluation. For example, a task that every model solves may be replaced with a harder task, while inconsistent grading may motivate revising a rubric or programmatic check. The agent records what it changed and whether the targeted metrics improved.

At the end of each round, the harness copies the current task data and verifier into an iteration directory (\texttt{iterations/iter\_N/}), checks task and response coverage, and runs the evaluator on the saved version. The resulting report and the agent's revision summary are stored alongside both artifacts. Before the next revision, the agent reviews this history and can restore an earlier task--verifier pair if a change performs worse. Saving both artifacts together preserves the correspondence between task definitions and scoring logic.

\paragraph{Time budget.}
We allocate a total wall-clock budget of 16 hours to each construction session. Task generation, model queries, verifier development, evaluation, and finalization share this budget. Before each round, the agent receives the remaining time and a summary of previous rounds, including their duration and evaluation results. It determines how long to spend on each revision and how many rounds to attempt. Refinement can end early when the agent chooses to submit or a configured stagnation criterion based on ranking agreement and score spread is met. Post-submission evaluation is performed separately from the construction budget.

\paragraph{Final submission.}
When refinement ends with time remaining, the harness launches a dedicated finalization round within the same budget. The agent reviews the saved iterations and their visible-model results, then chooses one of two submission strategies. It can restore the complete task set and verifier from a single iteration, or select useful tasks from multiple iterations and assemble a verifier that supports every retained task. In the latter case, the agent is instructed to remove duplicate tasks and reconcile their scoring logic.

The agent places the selected task data and verifier in the workspace root as its final submission. This round focuses on selecting and consolidating existing work. If the budget is exhausted before a finalization round can run, the current working task data and verifier are used. The agent is therefore instructed to keep its best available benchmark in the working files throughout construction. The harness evaluates the submitted pair; it does not automatically select the highest-scoring snapshot.

\subsection{Construction Safeguards}
\label{sec:method:anticheat}

We use construction rules and independent evaluation to reduce opportunities for gaming. The agent is instructed to create original tasks and to leave the supplied reference scores and evaluation script unchanged. The reference material supplied to its workspace excludes held-out models. The model-query proxy exposed to the construction agent also checks requests against a visible-model allowlist and rejects requests for other model names. The agent's response records are used for development feedback. After submission, the harness reissues the final task prompts to every visible and held-out model and recomputes scores using the submitted verifier. This removes agent-written response records from the final score calculation and provides a check against fabricated or selectively recorded answers. Held-out evaluation additionally tests whether the benchmark's comparisons extend beyond the models used to guide construction. These checks target response manipulation and overfitting to visible candidates; they do not provide a complete audit of arbitrary verifier code.

\subsection{Evaluation Protocol}
\label{sec:method:evaluation}

\paragraph{Independent evaluation.}
After construction, the harness independently collects new responses from all visible and held-out models and scores them with the submitted verifier. All final metrics are computed from this pass. We assess whether the generated benchmark preserves model rankings, whether this agreement extends to held-out models, and whether its scores retain the reference scale and separation. Each candidate model $m$ has two scores: $r_m$ is its score on the original reference benchmark, and $\hat r_m$ is its score on the generated benchmark, obtained by averaging verifier scores as in Equation~\eqref{eq:benchmark-score}. The hat thus distinguishes the generated-benchmark score from the reference score; neither symbol denotes a rank. We write $\mathcal{M}$ for the evaluated model set and $K=|\mathcal{M}|$ for its size.

\paragraph{Ranking fidelity.}
Pairwise accuracy asks a simple question: when the reference benchmark ranks one model above another, does the generated benchmark agree? For two models $m$ and $n$, the reference score difference is $r_m-r_n$, and the generated-benchmark score difference is $\hat r_m-\hat r_n$. The pair is correctly ordered when these differences have the same sign. Let $\mathcal{P}$ contain all unordered pairs with distinct reference scores, counting each pair once. The fraction of correctly ordered pairs is
\begin{equation}
    \mathrm{PAcc}(\mathcal{P}) =
    \frac{1}{|\mathcal{P}|}
    \sum_{(m,n)\in\mathcal{P}}
    \mathbbm{1}\!\left[
        (\hat r_m-\hat r_n)(r_m-r_n)>0
    \right].
    \label{eq:pairwise}
\end{equation}
Here, $\mathbbm{1}[\cdot]$ equals one if the condition holds and zero otherwise: a positive product means that both benchmarks prefer the same model. Reference ties are excluded because they do not specify a strict ordering; a tie on the generated benchmark counts as incorrect for a strictly ordered reference pair. Higher pairwise accuracy indicates more consistent model comparisons.

Complementing these pairwise comparisons, Spearman correlation measures agreement across the full ranking. We first replace each score with its rank among the evaluated models: $a_m=\operatorname{rank}(\hat r_m)$ for the generated benchmark and $b_m=\operatorname{rank}(r_m)$ for the reference benchmark, using the same ranking convention and assigning average ranks to ties. We then compute the correlation between these two rank vectors:
\begin{equation}
    \rho =
    \frac{\sum_{m\in\mathcal{M}}(a_m-\bar a)(b_m-\bar b)}
    {\sqrt{\sum_{m\in\mathcal{M}}(a_m-\bar a)^2}
     \sqrt{\sum_{m\in\mathcal{M}}(b_m-\bar b)^2}},
    \label{eq:spearman}
\end{equation}
where $\bar a$ and $\bar b$ are the mean ranks across $\mathcal{M}$. Values closer to one indicate stronger agreement with the reference ranking.

\paragraph{Candidate generalization.}
To distinguish agreement on familiar candidates from generalization to new ones, we separate pairs of visible models from pairs involving at least one held-out model. The visible and held-out sets are $\mathcal{M}_{\mathrm{vis}}$ and $\mathcal{M}_{\mathrm{held}}$, respectively. Held-out reference scores and trajectories are withheld during construction, and the model-query proxy permits queries only to visible models. The corresponding pair sets are
\begin{equation}
    \begin{aligned}
        \mathcal{P}_{\mathrm{vis}}
        = \{(m,n)\in\mathcal{P}:
             m,n\in\mathcal{M}_{\mathrm{vis}}\}, \quad
        \mathcal{P}_{\mathrm{held}}
        = \mathcal{P}\setminus\mathcal{P}_{\mathrm{vis}}.
    \end{aligned}
    \label{eq:candidate-pairs}
\end{equation}
We report $\mathrm{PAcc}(\mathcal{P}_{\mathrm{vis}})$ for visible models and $\mathrm{PAcc}(\mathcal{P}_{\mathrm{held}})$ for generalization. The latter includes both visible--held-out and held-out--held-out comparisons. When reporting Spearman correlation for a model subset, ranks and correlation are computed within that subset.

\paragraph{Calibration and resolution.}
Correct rankings alone do not imply accurate scores: a benchmark can preserve the ordering while shifting or compressing all scores. We therefore also compare the score values directly. Mean absolute error (MAE) measures the average magnitude of the discrepancy, while signed bias measures its direction:
\begin{equation}
    \begin{aligned}
        \mathrm{MAE} = \frac{1}{K}\sum_{m\in\mathcal{M}}|\hat r_m-r_m|, \quad
        \mathrm{Bias} = \frac{1}{K}\sum_{m\in\mathcal{M}}(\hat r_m-r_m).
    \end{aligned}
    \label{eq:calibration}
\end{equation}
Lower MAE indicates closer score agreement; positive and negative bias indicate overestimation and underestimation, respectively. To measure how strongly the benchmark separates models, we report:
\begin{equation}
    \mathrm{Spread} =
    \frac{\sigma(\{\hat r_m\}_{m\in\mathcal{M}})}
         {\sigma(\{r_m\}_{m\in\mathcal{M}})},
    \label{eq:spread}
\end{equation}
where $\sigma(\{x_m\}) = \sqrt{K^{-1}\sum_{m\in\mathcal{M}}(x_m-\bar x)^2}$ and $\bar x=K^{-1}\sum_{m\in\mathcal{M}}x_m$. A spread ratio near one indicates a similar degree of score separation to the reference benchmark; values below or above one indicate compressed or exaggerated differences, respectively.

%% file: sections/experiment.tex
\section{Experiments}
\label{sec:experiments}

\subsection{Experimental Setup}

\paragraph{Reference benchmarks.}
We evaluate \method using SWE-bench Verified and Terminal-Bench 2.1 as reference benchmarks. SWE-bench Verified contains 500 human-screened repository issue-resolution tasks \citep{jimenez2024swe,openai2024swebenchverified}, and Terminal-Bench 2.1 includes 89 command-line tasks \citep{merrill2026terminal,terminalbench2026v21}. For each reference, the agent receives the visible models' aggregate scores and reference trajectories, as described in Section~\ref{sec:formulation}. The scores provide the construction targets, while the trajectories inform the design of new tasks.

\paragraph{Models.}
Each reference uses ten candidate models, split into seven visible and three held-out candidates. The visible candidates are MiniMax-M3~\citep{minimax2026m3}, DeepSeek-V4-Pro~\citep{xu2026deepseek}, Gemini-3.1-Pro and Gemini-3.5-Flash~\citep{google2026gemini3}, GLM-5.1 and GLM-5.2~\citep{zai2026glm51,zai2026glm52}, and Kimi-2.5~\citep{kimi2026k25}. The held-out candidates are Claude-Opus-4.7~\citep{anthropic2026opus47}, GPT-5.5~\citep{openai2026gpt55}, and Kimi-K2.6~\citep{kimi2026k26}. These candidate roles remain hidden from the construction agent during the generation of queries and their corresponding verifiers.

\paragraph{Experimental settings.}
We evaluate five construction agents on each reference benchmark: GLM-5.2~\citep{zai2026glm52}, DeepSeek-V4-Pro and DeepSeek-V4-Flash~\citep{xu2026deepseek}, GPT-5.4~\citep{openai2026gpt54}, and Claude-Opus-4.7~\citep{anthropic2026opus47}. Each session receives a 16-hour construction budget and a target of 50 tasks. The agent jointly revises its task data and verifier, with no fixed number of iterations. It may submit early, and the harness can stop refinement when progress stalls. Construction details and prompt to control the agent is put into Appendix~\ref{app:prompts}. During finalization, the agent selects a saved task--verifier pair or combines tasks from multiple rounds with a compatible verifier, following Section~\ref{sec:method:iteration}. The evaluated task counts range from 40 to 65, as reported in Table~\ref{tab:results}.

\paragraph{Evaluation metrics.}
We follow the independent confirmation protocol described in Section~\ref{sec:method:evaluation}. Specifically, we collect fresh responses from all candidate models and evaluate them using the submitted verifier. We report overall pairwise accuracy, visible pairwise accuracy, and held-out-involved pairwise accuracy, together with overall Spearman correlation, MAE, Bias, and Spread. Detailed definitions of these metrics are provided in Equations~\eqref{eq:pairwise}--\eqref{eq:spread}.

\begin{table*}[t]
\centering
\small
\setlength{\tabcolsep}{6pt}
\renewcommand{\arraystretch}{1.08}
\caption{Results from confirmed evaluations.
All, Visible, and Held-out report pairwise accuracy on
$\mathcal{P}$, $\mathcal{P}_{\mathrm{vis}}$, and
$\mathcal{P}_{\mathrm{held}}$, respectively. $N$ is the number of tasks evaluated; $\rho$, MAE, Bias, and Spread
are computed across all ten candidates.
Bold marks the best value in each metric column within each reference,
including ties; Bias and Spread are best at zero and one, respectively.}
\label{tab:results}
\begin{tabular}{@{}lrrrrrrrr@{}}
\toprule
Construction agent & $N$ & \multicolumn{3}{c}{Pairwise accuracy $\uparrow$}
& $\rho\uparrow$ & MAE$\downarrow$ & Bias$\to0$ & Spread$\to1$ \\
\cmidrule(lr){3-5}
& & All & Visible & Held-out & & & & \\
\midrule
\multicolumn{9}{@{}l}{\textbf{Terminal-Bench 2.1}} \\
GLM-5.2 & 50 & 0.489 & 0.381 & 0.583 & 0.042 & 0.277 & $+0.277$ & 0.263 \\
DeepSeek-V4-Pro & 50 & 0.622 & \textbf{0.667} & 0.583 & 0.297 & \textbf{0.153} & $\mathbf{+0.134}$ & \textbf{0.877} \\
DeepSeek-V4-Flash & 65 & 0.444 & 0.571 & 0.333 & $-0.176$ & 0.336 & $+0.336$ & 0.121 \\
GPT-5.4 & 50 & 0.467 & 0.524 & 0.417 & $-0.055$ & 0.158 & $+0.158$ & 0.057 \\
Claude-Opus-4.7 & 40 & \textbf{0.689} & 0.619 & \textbf{0.750} & \textbf{0.467} & 0.207 & $-0.207$ & 0.641 \\
\addlinespace[3pt]
\multicolumn{9}{@{}l}{\textbf{SWE-bench Verified}} \\
GLM-5.2 & 50 & 0.571 & 0.550 & 0.591 & 0.190 & 0.082 & $-0.060$ & \textbf{0.985} \\
DeepSeek-V4-Pro & 50 & 0.405 & \textbf{0.700} & 0.136 & $-0.193$ & 0.187 & $-0.173$ & 3.906 \\
DeepSeek-V4-Flash & 50 & 0.429 & 0.550 & 0.318 & 0.137 & \textbf{0.030} & $\mathbf{+0.018}$ & 0.428 \\
GPT-5.4 & 41 & \textbf{0.619} & 0.550 & \textbf{0.682} & 0.288 & 0.079 & $+0.079$ & 0.202 \\
Claude-Opus-4.7 & 50 & 0.595 & 0.500 & \textbf{0.682} & \textbf{0.588} & 0.222 & $+0.222$ & 0.265 \\
\bottomrule
\end{tabular}
\end{table*}

\subsection{Main Results}
We conduct experiments with a range of frontier models to investigate their performance and behavior in constructing efficient benchmarks. The corresponding results are reported in Table~\ref{tab:results}.
\paragraph{Ranking Fidelity}  We firstly analyze the ranking fidelity by treating different models as resaerch agent. Specifically, on Terminal-Bench 2.1, Claude-Opus-4.7 achieves the highest overall pairwise accuracy and Spearman correlation. On SWE-bench Verified, GPT-5.4 achieves the highest pairwise accuracy, while Claude-Opus-4.7 obtains the highest Spearman correlation. These results are consistent with external evaluations showing strong coding and research capabilities for GPT-5.4 and Claude-Opus-4.7. This suggests that the ability to construct effective and efficient benchmarks is closely related to the underlying coding and research.

\paragraph{Calibration and Resolution}

Moreover, we observe that six of the ten generated benchmarks have a spread ratio below 0.5, indicating compressed differences between model scores. A small MAE does not necessarily imply an accurate ranking either: DeepSeek-V4-Flash on SWE-bench achieves the lowest MAE, but its overall pairwise accuracy is only 0.4286. At the other extreme, DeepSeek-V4-Pro on SWE-bench has a spread ratio of 3.9056 and negative Spearman correlation, indicating exaggerated score differences alongside poor ranking agreement. GLM-5.2 on the same reference has a spread ratio close to one (0.9846), but only moderate pairwise accuracy (0.5714). These examples further demonstrate the comprehensiveness of our evaluation framework and highlight why ranking fidelity, calibration, and resolution should be assessed separately.

Overall, frontier models show promising but still limited ability to construct reliable benchmarks. Claude-Opus-4.7 demonstrates the strongest overall ranking fidelity, while GPT-5.4 achieves the highest pairwise accuracy on SWE-bench. However, the generated benchmarks remain far from being reliable substitutes for costly evaluations, particularly when performance is assessed on held-out models. Moreover, several models either compress score differences or exaggerate them without preserving the correct ranking, indicating that calibration alone is insufficient for reliable benchmarks. These results highlight the importance of carefully designed constraints in \method: held-out models and fresh responses should be reserved for independent validation, and generated benchmarks should be assessed on various metrics to prevent overfitting to visible examples or a single metric.

\begin{figure}[t]
\vspace{-0.2em}
    \centering
    \includegraphics[width=\linewidth]
        {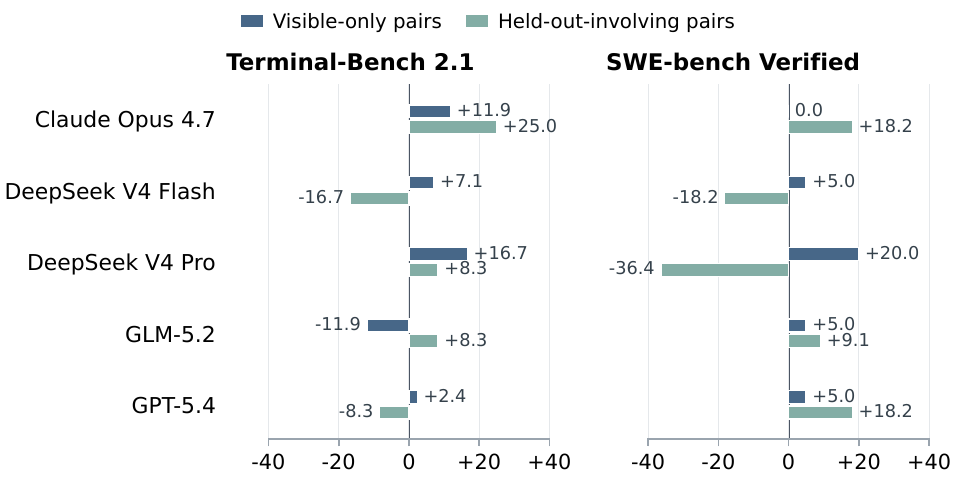}
    \vspace{-1.8em}
    \caption{Pairwise accuracy relative to random ordering for each result(accuracy minus 50\%). The zero line denotes chance performance. Visible-only bars show visible–visible pairs; held-out bars aggregate visible–held-out and held-out–held-out pairs by total correct and total comparisons.}
    \label{fig:generalization}
\end{figure}

\subsection{Candidate Generalization}

Figure~\ref{fig:generalization} compares visible and held-out-involved pairwise accuracy using the pair sets defined in Equation~\eqref{eq:candidate-pairs}. DeepSeek-V4-Pro on SWE-bench Verified illustrates a large gap: its visible accuracy is 0.7000, but held-out-involved accuracy falls to 0.1364. Thus, strong agreement on the models used during construction may not transfer to new candidates. Other runs retain stronger performance on comparisons involving held-out models. Claude-Opus-4.7 reaches 0.7500 held-out-involved accuracy on Terminal-Bench, while GPT-5.4 and Claude-Opus-4.7 both reach 0.6818 on SWE-bench. This metric includes comparisons between visible and held-out models as well as comparisons within the held-out set. As a supplementary diagnostic, the SWE-bench Claude-Opus-4.7 run scores 0/3 on held-out-only pairs because its three held-out scores are tied. Its higher pooled accuracy therefore comes from comparisons across the visible and held-out sets.

% \subsection{Phase of Evolution}

% \begin{figure}[t]
%     \centering
%     \includegraphics[width=\linewidth]
%         {figures/evolution_combined.pdf}
%     \caption{Evolution of benchmark quality on the visible model set during construction.}
%     \label{fig:phaseofevolution}
% \end{figure} 

% We further investigate how benchmark quality evolves over time by tracking pairwise accuracy and Pearson correlation throughout the benchmark development process. This analysis provides insights into the agents' refinement behavior and how the quality of generated benchmarks changes with increasing development time. The corresponding results are shown in Figure~\ref{fig:phaseofevolution}. Specifically,  Figure~\ref{fig:phaseofevolution} illustrates several limitations in how agents manage construction. On SWE-bench Verified, Claude-Opus-4.7 reaches a visible-set pairwise accuracy of 0.80 after approximately 10.4 hours, but continues until 13.1 hours without further improvement. On Terminal-Bench, DeepSeek-V4-Flash produces only two iteration snapshots over the 16-hour budget, with pairwise accuracy remaining at 0.286. The confirmation results also suggest possible overfitting: DeepSeek-V4-Flash reaches 0.90 visible-set pairwise accuracy on SWE-bench during construction, but achieves only 0.55 on the same visible models under independent confirmation. These cases suggest that agents may benefit from guidance on when to stop, how to distribute time across revisions, and how to check whether apparent improvements survive fresh response collection in the future.

\subsection{Efficiency Analysis and Case Study}

\begin{figure}[t]
\vspace{-1.5em}
    \centering
    \includegraphics[width=\linewidth]
        {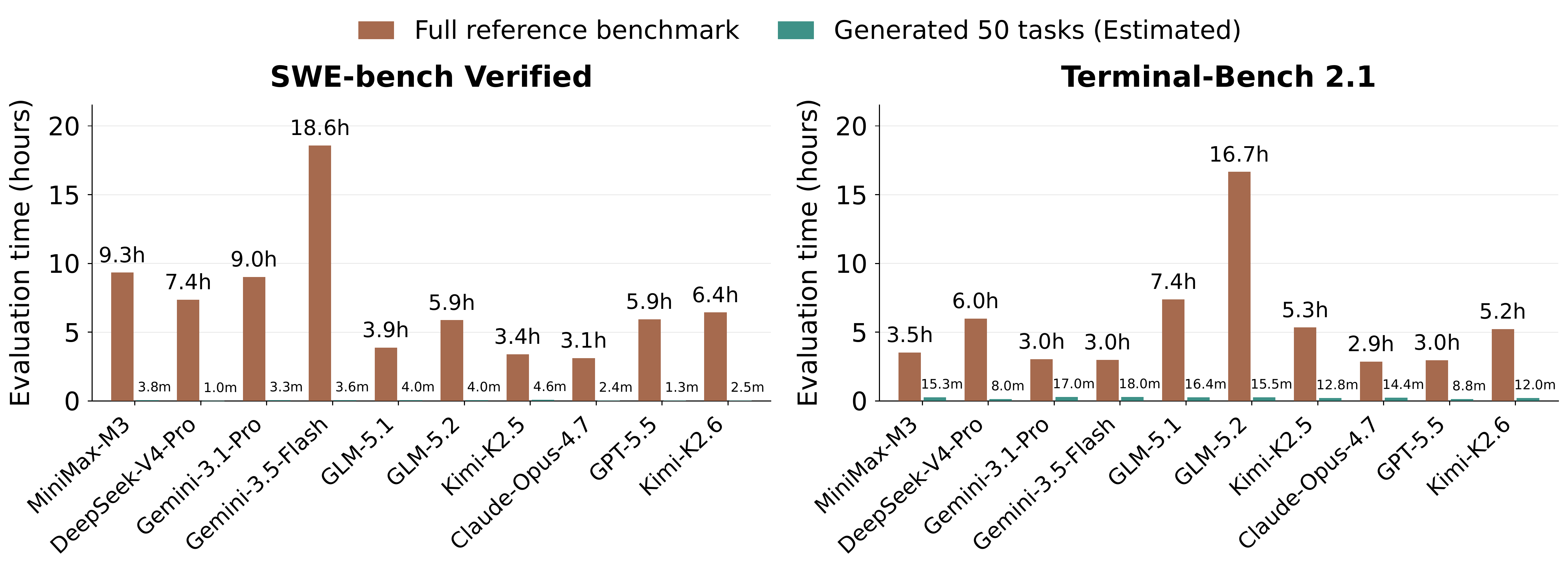}
    \vspace{-2.1em}
    \caption{Evaluation Time of Agent Benchmarks and Generated Lightweight Benchmarks.}
    \label{fig:time-efficiency}
\end{figure} 

\begin{figure}[t]
    \centering
    \includegraphics[width=\linewidth]
        {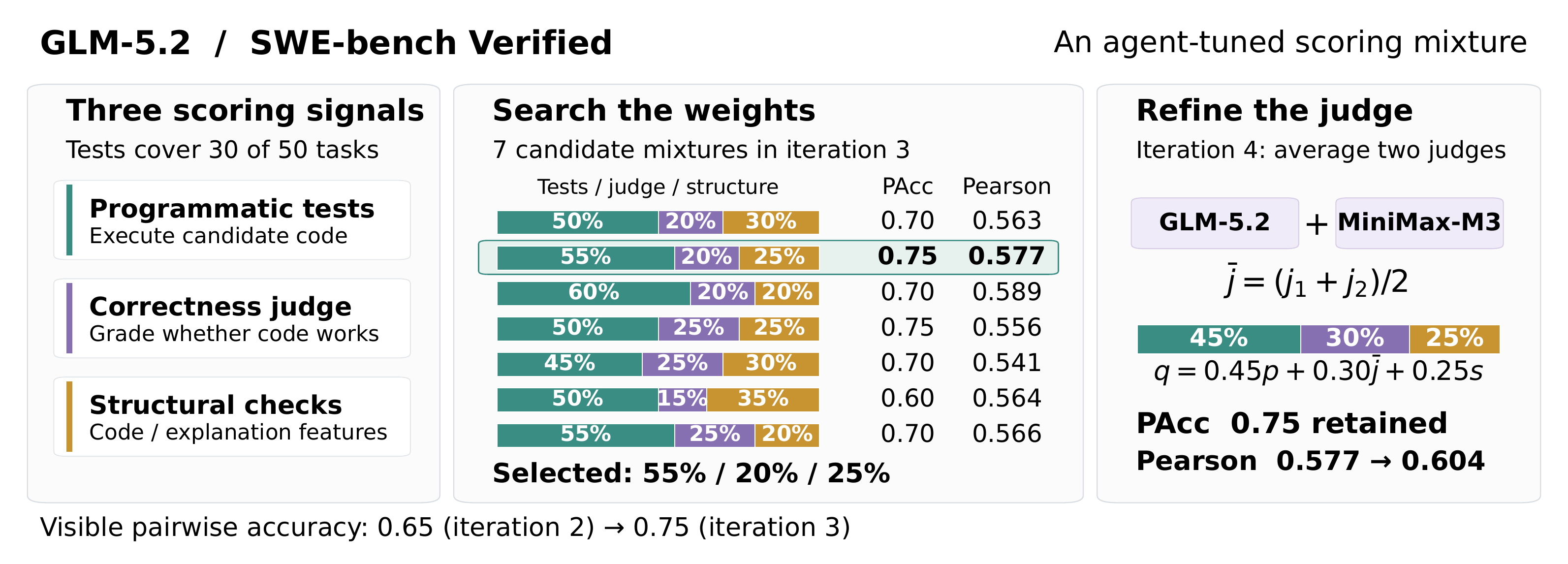}
    \vspace{-2.1em}
    \caption{Case Study of Model's Strategy to improve performance of  building efficient benchmarks.}
    \label{fig:case-study-weights}
\end{figure} 

We report the evaluation time for the reference benchmarks, which require cumbersome evaluation procedures, alongside the evaluation time for the benchmarks generated by our agent, as shown in Figure~\ref{fig:time-efficiency}. The efficiency comparison suggests substantial potential to reduce inference time during model evaluation. Compact generated benchmarks could make repeated evaluation much faster, although the estimated savings remain to be validated through direct timing measurements. We further analyze the model's behavior during the evolution process by analyzing their trajectories and strategies they use. Figure~\ref{fig:case-study-weights} illustrates an agent adopting a familiar idea from classical ensemble methods: weighting complementary signals according to evaluation feedback. GLM-5.2 combines programmatic tests, correctness judgments, and structural checks, searches seven  weight combinations, and selects a 55/20/25 mixture. It subsequently averages two judges and retunes the weights to 45/30/25, retaining visible pairwise accuracy of 0.75 while improving Pearson correlation from 0.577 to 0.604. This example suggests that agents can autonomously construct evaluators by adapting established ideas such as weighted aggregation and ensemble scoring. More results are shown in Appendix~\ref{app:case-studies} and \ref{app:task-examples}.

% \subsection{Independent Response Confirmation}

% The Terminal-Bench DeepSeek-V4-Pro record provides a comparison before
% and after independent confirmation. The earlier evaluation combined
% agent-collected visible responses with harness-collected held-out
% responses, yielding overall pairwise accuracy of 0.800 and
% held-out-involved accuracy of 0.708. Recollecting responses for all
% candidates changed these values to 0.622 and 0.583, respectively.
% This example shows that response recollection can materially change the
% measured agreement. All main results therefore use the confirmation
% pass described in Section~\ref{sec:method:evaluation}.

% \subsection{Benchmark Size and Cost}

% The evaluated benchmarks contain 40--65 tasks. Their scoring procedures
% may include model-based judges or code execution, so task count alone
% does not establish an efficiency gain. The available confirmation
% summaries do not include complete raw response and verifier artifacts
% or matched token, runtime, API, and compute logs. We therefore report
% benchmark size without claiming a measured cost reduction relative to
% the reference evaluations.

%% file: sections/conclusion.tex
\section{Conclusion}

In this work, we study automatic benchmark development as an emerging capability of LLM research agents. We formulate a setting in which agents construct efficient proxy benchmarks from expensive reference evaluations and introduce \method, a systematic framework for assessing the quality of generated benchmarks along fidelity, efficiency, and robustness. Our evaluation considers whether generated benchmarks preserve model rankings, generalize to held-out models and fresh responses, and provide informative scores while reducing evaluation cost. Beyond measuring benchmark quality, our framework also reveals common failure modes of automatically generated evaluators and provides a practical environment for iterative benchmark refinement. We hope this work provides a step toward reducing the human effort required for evaluation design and, more broadly, toward increasingly automated model-development pipelines in which agents can not only solve tasks, but also help construct reliable mechanisms for evaluating progress.

%% file: sections/appendix.tex
% Self-contained appendix: place after the bibliography.
% Uses the paper's existing amsmath, booktabs, and graphicx packages.
\appendix

\section{Agent Prompts and Construction Details}
\label{app:prompts}

This appendix expands the construction setting with the prompt structure,
artifact format, and examples of how agents use feedback. We then examine
saved construction trajectories to connect the aggregate results to concrete
changes in tasks and scoring code. The cases are selected to illustrate
different behaviors, rather than to estimate their frequency.

\subsection{What the Agent Receives}

The initial prompt defines the reference benchmark, the visible candidate
models, a target of 50 tasks, and a shared construction budget of 960 minutes.
The agent receives the visible candidates' aggregate reference scores and
reference trajectories. The aggregate scores are the numerical targets for
benchmark construction; the trajectories provide examples of the original
tasks, model behavior, and success or failure outcomes. Held-out candidates'
reference scores and trajectories are excluded from these construction
inputs.

The workspace separates the editable benchmark from the supplied reference
material. The agent creates and revises \texttt{responses.json}, which
contains the task definitions and collected answers, and \texttt{verify.py},
which assigns scores. It may write collection scripts, draft task pools,
and scoring experiments in \texttt{tmp\_operations/}. The instructions ask
it to preserve \texttt{truth.json}, \texttt{evaluate.py}, and the reference
trajectories. Completed iterations retain their own task data, verifier,
evaluation report, and revision summary.

The prompt encourages the agent to inspect selected reference trajectories
rather than reading every large file. It asks for new tasks spanning task
types, domains, and difficulty levels, and explicitly discourages copying
or lightly rewording reference instances. This gives agents substantial
freedom over what their proxy benchmark measures while keeping the
reference ranking fixed.

\subsection{Prompt Structure}
\label{app:prompt-templates}

There are three prompt stages: initialization, refinement, and finalization.
The blocks below are condensed descriptions of the repository's prompt
templates, not verbatim transcripts of every historical run. Run-specific
model names, time values, and history digests are represented by
placeholders. Transport details and repeated workspace reminders are
omitted.

\begin{quote}
\small
\textbf{Initialization prompt (condensed).}

You are constructing a new benchmark for \emph{[reference benchmark]}.
Design approximately 50 original tasks and an executable verifier that
rank the visible candidate models similarly to their reference results.
The total construction budget is 960 minutes; \emph{[remaining time]}
minutes remain.

Read the supplied reference scores and, when useful, inspect reference
trajectories to identify relevant challenges. Do not copy the reference
tasks. Include variation in task type, domain, and difficulty, and avoid
near-duplicate questions.

Create the task data, collect actual responses from every visible
candidate, and implement the verifier. You may revise prompts, reference
answers, rubrics, tests, and scoring rules. Preserve the supplied reference
scores and evaluator. Keep helper scripts and draft data in the scratch
directory.

Run the evaluator, inspect ranking agreement, bias, spread, and missing
responses, then write an iteration summary. Leave a working task--verifier
pair in the workspace root. Save enough time for later revisions and final
submission.
\end{quote}

The first prompt also specifies the response schema and verifier interface.
It explains that missing answers can distort candidate rankings and asks
the agent to retry collection failures rather than interpreting them as
evidence of task difficulty. It warns that final evaluation includes
candidates unavailable during construction.

\begin{quote}
\small
\textbf{Refinement prompt (condensed).}

Approximately \emph{[remaining time]} minutes remain across all remaining
rounds. Review the saved summaries and reports before editing.
\emph{[History digest: round durations, validity, ranking metrics, bias,
and spread.]}

First diagnose a concrete problem. Identify incorrectly ordered candidate
pairs, tasks on which every model receives the same score, overly lenient
or harsh scoring, missing responses, or unreliable task-specific grading.
Read representative answers and the corresponding verifier logic.

Then make changes that address this diagnosis. You may replace tasks,
revise tests or rubrics, repair response coverage, or restore an earlier
task--verifier pair. Re-run the evaluator and record whether the targeted
problem improved. If a revision regresses, keep the best working version
in the workspace root.

Write what changed, why, the observed result, and any rollback.
Set \texttt{ready\_to\_submit=true} when the current benchmark is usable and
further revisions are unlikely to help. The harness checks validity before
honoring early submission.
\end{quote}

Refinement uses the continuing agent session and the persistent workspace.
The supplied history includes round-level timing and a pacing summary,
such as average time per completed round. These are planning aids: there
is no fixed duration assigned to an ordinary iteration. The agent decides
when to end its turn and how much work to attempt before the shared budget
expires.

\begin{quote}
\small
\textbf{Finalization prompt (condensed).}

This is the final round. Approximately \emph{[remaining time]} minutes
remain. Submit a complete benchmark using the task data and verifier in
the workspace root.

Choose one saved iteration and restore both artifacts, or combine useful
tasks from several iterations and assemble a verifier that supports all
retained tasks. Preserve unique task identifiers and remove duplicate
tasks. Copying task data alone is insufficient if its scoring logic is
missing.

Prioritize a working verifier and complete task records. If time permits,
run one final check and briefly record which version or combination was
selected. Focus on consolidating existing work. The harness evaluates the
root artifacts; it does not select the best snapshot on the agent's behalf.
\end{quote}

Finalization is a separate instruction because selecting a submission is
a different decision from proposing another experiment. It is launched
only when construction ends with enough time remaining. A round that
consumes the entire budget can therefore leave the current root artifacts
as the submission without a dedicated finalization turn.

\subsection{Task Records and Verifier Interface}
\label{app:artifact-format}

A task record contains a unique identifier, its prompt, an optional
reference answer or rubric, and the responses collected during
construction. The following is a structural example:
\begin{quote}
\small
\begin{verbatim}
[
  {
    "task_id": "<unique identifier>",
    "prompt": "<new task prompt>",
    "reference_answer": "<answer or rubric>",
    "model_responses": {
      "<visible model>": "<raw response>"
    }
  }
]
\end{verbatim}
\end{quote}
The response mapping contains an entry for each visible candidate.
The record is both a task definition and a development-time response
container; independently collected answers replace the development
responses for final evaluation.

The verifier implements
\begin{quote}
\small
\begin{verbatim}
verify(response, answer, task) -> float
\end{verbatim}
\end{quote}
with scores in $[0,1]$. The evaluator supplies the response, the
agent-authored reference answer, and task metadata. In the inspected
implementation, this metadata includes the task identifier, prompt,
reference answer, and candidate name. Agents consequently implement both
task-ID dispatch and reference-answer-based dispatch. For example,
some store a JSON-encoded specification in the reference answer,
including a function name, test inputs, expected outputs, and grading
category.

The task schema does not by itself guarantee meaningful scoring.
An executable verifier may recognize only some task identifiers, ignore
a reference answer, or return a constant on unfamiliar tasks. Likewise,
a verifier can contain cached judgments or model-dependent branches.
These are properties of the submitted scoring code that require inspection
beyond checking whether the file executes.

\subsection{Revision Records and Submission Checks}
\label{app:submission-details}

The requested revision summary records the diagnosis,
\texttt{what\_changed}, \texttt{why}, \texttt{compared\_to\_previous},
\texttt{intermediate\_validation}, \texttt{did\_it\_work},
\texttt{rolled\_back\_to}, \texttt{ready\_to\_submit}, and the agent's
final metric report. The harness supplements the saved record with its
own validity decision, evaluation results, and timing. An automatically
generated summary indicates that the agent did not provide the requested
write-up; it is not evidence that the agent deliberately chose to keep
the current version.

The inspected validity check requires at least 80\% of the requested task
count, giving a minimum of 40 tasks for a 50-task target. It also requires
at least 80\% of tasks to contain responses for every visible candidate,
an edited verifier rather than the original placeholder, and an evaluation
that returns a non-null Spearman correlation. These checks determine
whether a submission can be evaluated. They do not verify that every task
has an appropriate scoring rule.

The default stagnation rule uses a score
$s=\rho-0.25\min(1,|\mathrm{Spread}-1|)$ to track progress between rounds.
An increase must exceed 0.01 to reset the counter, and three consecutive
non-improving rounds trigger early stopping. This is a construction
heuristic, separate from the evaluation dimensions reported in the paper.
Agents can also request early submission, subject to the validity check.
The recorded stopping decision therefore depends on both agent behavior
and harness rules.

\section{Construction Histories}
\label{app:histories}

Table~\ref{tab:appendix-run-inventory} summarizes the exact runs used in
the reported experiments. Round counts include unsuccessful iterations:
a rejected task set still consumes construction time and may influence
the next revision. Finalization is counted separately because its
instruction is to select or consolidate existing artifacts.

\begin{table*}[t]
\centering\small
\setlength{\tabcolsep}{6pt}
\caption{Construction histories for the selected runs. Iter./Final. counts refinement and finalization rounds; Invalid lists rejected saved rounds. $N$ is the independently confirmed task count. Times include recorded construction checks.}
\label{tab:appendix-run-inventory}
\begin{tabular}{@{}lcccrl@{}}
\toprule
Construction agent & Iter./Final. & Invalid & $N$ & Time & Stop \\
\midrule
\multicolumn{6}{@{}l}{\textbf{SWE-bench Verified}} \\
GLM-5.2 & 6/1 & 1 & 50 & 8h 07m & Agent submit \\
DeepSeek-V4-Pro & 10/1 & 1 & 50 & 4h 33m & Agent submit \\
DeepSeek-V4-Flash & 4/1 & 2 & 50 & 5h 37m & Agent submit \\
GPT-5.4 & 6/1 & 1 & 41 & 7h 00m & Agent submit \\
Claude-Opus-4.7 & 10/1 & -- & 50 & 13h 07m & Agent submit \\
\addlinespace[3pt]
\multicolumn{6}{@{}l}{\textbf{Terminal-Bench 2.1}} \\
GLM-5.2 & 2/0 & -- & 50 & $\geq$6h 21m & Agent Submit \\
DeepSeek-V4-Pro & 5/1 & 2 & 50 & 7h 09m & Agent submit \\
DeepSeek-V4-Flash & 2/0 & -- & 65 & 16h 00m & Timeout \\
GPT-5.4 & 3/1 & 1, 2, 3 & 43 & 5h 11m & Agent submit \\
Claude-Opus-4.7 & 10/1 & 5 & 40 & 11h 46m & Agent submit \\
\bottomrule
\end{tabular}
\end{table*}

\subsection{Timing and Stopping Behavior}

Elapsed time is reconstructed from the harness's remaining-budget
counters, round durations, and in-session evaluation times. It includes
recorded finalization and construction checks, while subsequent
independent confirmation is excluded. The counters are rounded, so the
times should be interpreted at the scale of minutes.

The runs exhibit different stopping patterns. On SWE-bench Verified,
DeepSeek-V4-Pro ends after approximately 4 hours 33 minutes following
the stagnation rule, while Claude-Opus-4.7 submits after approximately
13 hours 7 minutes. On Terminal-Bench, DeepSeek-V4-Pro requests early
submission and finishes finalization after approximately 7 hours
9 minutes. DeepSeek-V4-Flash reaches the 16-hour limit during its second
iteration.

\begin{figure*}[t]
\centering
\includegraphics[width=\textwidth]
  {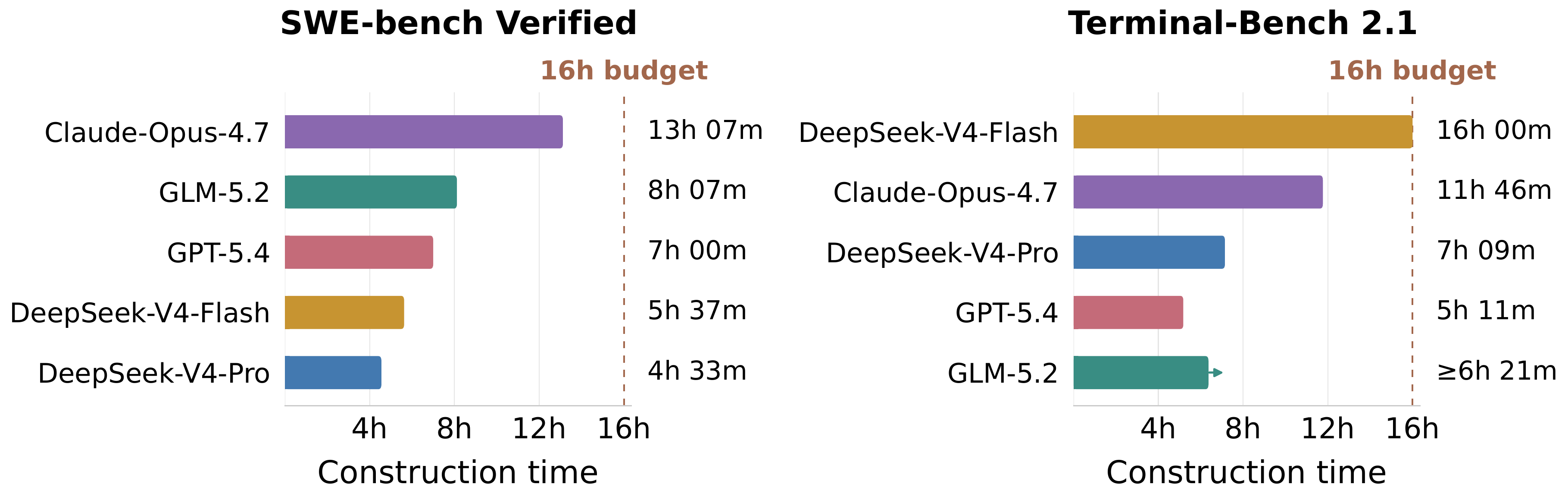}
\caption{Construction time for the selected runs. Dashed lines mark the
16-hour budget. The arrow for GLM-5.2 on Terminal-Bench denotes a recorded
lower bound because its final timing is unavailable.}
\label{fig:appendix-budget}
\end{figure*}

The Terminal-Bench GLM-5.2 log contains a manually completed second-round
record without an elapsed time. Its first completed round establishes
a lower bound of approximately 6.37 hours, but the available counters
do not establish the total session duration. Figure~\ref{fig:appendix-budget}
marks this value as a lower bound rather than assigning an estimated end
time. Each bar represents one selected run, so the figure reports neither
an average across repetitions nor an error bar.

\section{Scoring Strategies and Trajectory Case Studies}
\label{app:case-studies}

\subsection{GLM-5.2: Designing a Scoring Mixture and Searching Its Weights}
\label{app:case-glm-weights}

The SWE-bench GLM-5.2 run provides an example of an agent developing a
scoring strategy beyond writing individual tasks. By iteration 2, the
agent combines programmatic test results with structural checks.
Its next diagnosis is that programmatic tests cover only 30 of the
50 tasks, while structural heuristics give little separation on the
remaining tasks. It also observes that an earlier general-quality judge
was too lenient toward verbose responses. Rather than abandoning
model-based judging, it changes the judge instruction to focus on
whether the proposed code actually works.

\paragraph{Complementary scoring signals.}
The agent constructs three raw signals: programmatic correctness $p$,
a correctness-focused model judgment $j$, and structural checks $s$.
The last component uses heuristic features of the code and explanation,
rather than an independent execution test. The agent combines these
signals as
\begin{equation}
\begin{aligned}
q &= w_p p+w_j j+w_s s,\\
w_p+w_j+w_s &= 1.
\end{aligned}
\label{eq:appendix-score-mixture}
\end{equation}
Here $q$ is the raw component mixture, before the run's subsequent score
calibration. The inspected scoring script uses the structural score as
a fallback when a programmatic score is unavailable. Consequently,
the mixture can still produce a score for tasks without executable tests.

\paragraph{Searching the weights.}
Iteration 3 evaluates seven combinations against the visible reference
scores, as recorded in the agent's revision summary
(Table~\ref{tab:weight-search}). The selected combination is
$w_p=0.55$, $w_j=0.20$, and $w_s=0.25$.
Two candidates attain pairwise accuracy 0.75; the selected mixture has
the higher Pearson correlation among these two.
This is empirical weight selection, rather than a reported
gradient-based regression procedure.

\begin{table}[t]
\centering\small
\setlength{\tabcolsep}{5pt}
\caption{Weight combinations tried by GLM-5.2 in iteration 3, as reported
in its revision summary. Weights are percentages. Bold marks the selected
mixture; these are visible-set development results.}
\label{tab:weight-search}
\begin{tabular}{@{}rrrrr@{}}
\toprule
$w_p$ & $w_j$ & $w_s$ & PAcc & Pearson \\
\midrule
50 & 20 & 30 & 0.70 & 0.563 \\
\textbf{55} & \textbf{20} & \textbf{25} & \textbf{0.75} & \textbf{0.577} \\
60 & 20 & 20 & 0.70 & 0.589 \\
50 & 25 & 25 & 0.75 & 0.556 \\
45 & 25 & 30 & 0.70 & 0.541 \\
50 & 15 & 35 & 0.60 & 0.564 \\
55 & 25 & 20 & 0.70 & 0.566 \\
\bottomrule
\end{tabular}
\end{table}

The original harness checkpoints show visible pairwise accuracy rising
from 0.65 in iteration 2 to 0.75 in iteration 3.
The agent also repairs several truncated responses during this round,
so this comparison describes the complete revision, not an isolated
effect of changing weights. The useful design behavior is that the
agent identifies missing evaluation evidence, adds a complementary
signal, and explicitly searches how much that signal should contribute.

\paragraph{Refining the judge ensemble.}
In iteration 4, the agent examines disagreements between its correctness
judge and execution tests. Its summary highlights, for example, a tree
serialization answer receiving a perfect judge score despite failing
the programmatic checks. It adds MiniMax-M3 as a second judge alongside
GLM-5.2 and averages their scores:
\begin{equation}
\begin{aligned}
\bar j &= \tfrac{1}{2}(j_{\mathrm{GLM}}+j_{\mathrm{MiniMax}}),\\
q &= 0.45p+0.30\bar j+0.25s.
\end{aligned}
\label{eq:appendix-judge-ensemble}
\end{equation}
The revision summary reports eight combinations with the averaged judge.
The 45/30/25 mixture retains visible pairwise accuracy 0.75 while
Pearson correlation increases from 0.577 to 0.604. The agent also tries
using the second judge alone: its reported best Pearson value is higher,
but pairwise accuracy falls to 0.65, so that alternative is not retained.
This makes the selection criterion concrete: the agent seeks better
score agreement while preserving model-ordering decisions.

\paragraph{Implementation and interpretation.}
The retained score-construction helper explicitly averages the two judge
outputs and computes the 45/30/25 combination. It then applies additional
scaling before saving scores. Figure~\ref{fig:case-study-weights} therefore
depicts the learned scoring recipe at the component level, not every
operation in the submitted verifier.

The mixture search is a constructive behavior observed during development.
It does not establish that these particular weights transfer unchanged
to fresh responses or held-out candidates. Later rounds modify scoring
artifacts, and the final independently confirmed GLM-5.2 result has overall
pairwise accuracy 0.5714. The example's contribution is the agent's
ability to formulate and compare evaluation strategies; the final
confirmation remains the assessment of the submitted benchmark.

\subsection{GPT-5.4: Fitting Visible Scores Through Surface Features}
\label{app:case-gpt54}

The construction run uses high reasoning effort to produce a
software-engineering task set. It receives SWE-bench Verified reference
scores during construction; the later Terminal-Bench comparison uses a
different reference. Its main revisions retain the task set while
adjusting response collection and a deterministic verifier based on
task-specific keywords, code and test markers, and response length.
The discussion here describes construction and diagnostic trends;
final numerical results are reported separately.

\paragraph{Visible feedback and scorer revisions.}
The initial snapshot leaves visible ranking errors. Subsequent revisions
remove an overly broad refusal pattern that also matches a valid JSX
\texttt{placeholder} attribute, recollect selected short Gemini responses,
and add a lexical penalty for a React request-ID pattern. These changes
improve agreement with the visible reference ordering. The agent then
adjusts the penalties for long and very long answers, further improving
the ordering on the saved responses.

Once the visible ordering stabilizes, the agent targets the score scale.
Later rounds fit an affine transformation to the visible predicted and
reference means and standard deviations, add a small bonus for high
keyword coverage at moderate response length, and recalibrate. The
resulting saved-response scores align more closely with the visible
reference scale: absolute error and systematic offset decrease, while
the score spread approaches that of the reference. These improvements
partly follow from explicitly fitting those properties and do not
independently establish transfer to new responses. The agent signals
readiness and submits well before exhausting its time budget.

\paragraph{Independent response collection.}
Re-scoring the same saved answers preserves the apparent fit, but does
not test its dependence on response collection. The agent's collection
scripts concatenate thinking and final-text blocks, and a recollection
helper retains a new answer only when it is at least as long as the
previous one. Some saved Flash answers are visibly incomplete; one
implementation stops in the middle of a conditional expression.

The independent pass freezes the prompts and verifier, takes only
complete final text, and retries failures or truncation without selecting
answers by score or length. It uses a larger initial output budget and
increases the limit when truncation occurs. Consequently, score changes
reflect a changed collection protocol as well as response variability;
they are not, by themselves, evidence of deliberate fabrication.

On the matched subset of visible models with complete fresh responses,
agreement with the SWE-bench reference ordering declines, absolute error
increases, and the predicted scores become much more tightly clustered.
Flash's mean score rises toward the other candidates, removing much of
the apparent separation present in the original collection. The strong
fit to saved answers therefore does not persist under fresh collection.

\paragraph{Scoring sensitivity and remaining coverage.}
A negative control concatenates each task's keywords, a function
containing only \texttt{pass}, test-related words, and irrelevant padding.
These non-solutions still receive high scores. Renaming the model
identifier does not change the tested scores, but this rules out only
that tested form of identity dependence: the verifier still rewards
surface features without validating solutions.

A small number of visible DeepSeek and Kimi queries remain incomplete
in the recollection. Separate fresh evaluations of the GPT and Kimi
candidates produce a relative ordering that disagrees with their
SWE-bench reference ordering. Kimi-K2.6 was also the actual backend for
the training label \texttt{kimi-k2.5}, so it is not a model-unseen
held-out candidate in this run.

Combining successful-response means with the externally reported Opus
aggregate for a Terminal-Bench comparison likewise shows weak ranking
agreement and a compressed score spread. This diagnostic comparison is
provisional: visible response coverage is incomplete, the external Opus
result is available as an aggregate rather than an imported per-task
record, and construction used the SWE-bench reference. It should not be
interpreted as a fully confirmed Terminal-Bench construction result.

The case illustrates how visible feedback can improve calibration and
repair individual heuristics while leaving the scoring function sensitive
to collection artifacts and superficial answer features. Fresh-response
evaluation and negative controls reveal weaknesses that strong agreement
on the saved visible answers does not expose.

\subsection{DeepSeek-V4-Flash: A Narrow Core Selected Against Visible Scores}
\label{app:case-flash-swe}

On SWE-bench Verified, DeepSeek-V4-Flash begins with a 60-task snapshot
combining code execution, output prediction, and keyword-graded questions.
Its first valid visible pairwise accuracy is 0.60. The agent's diagnosis
is that many clean programming tasks give all visible models full credit,
leaving little variation to recover the reference ordering.

The next round collects a larger task pool and selects a nine-task subset
whose observed score patterns align with the visible reference scores.
Its summary reports improved ranking, but the harness rejects the
snapshot because it contains only nine tasks. These self-reported numbers
are therefore not plotted as a valid construction result.

Iteration 3 repairs missing responses and restores the minimum task count
by combining the nine-task core with 31 tasks described in the summary
as score-constant on the visible candidates. It also changes the score
calibration. The saved 40-task benchmark reaches visible pairwise accuracy
0.90 and Spearman correlation approximately 0.972. This is a useful
distinction between nominal benchmark size and the number of tasks
providing observed discrimination: meeting the count threshold does not
necessarily broaden the signal used to rank models.

In iteration 4, the agent explicitly identifies the narrow core as a
weakness. It reports trying additional task combinations but retaining
the current version when broader sets reduce visible agreement. The
40-task data and verifier are unchanged between the two saved snapshots.
The agent then requests early submission. Finalization produces a
50-task submission with the same visible pairwise accuracy of 0.90.

Under independently collected answers, the final benchmark achieves
visible pairwise accuracy 0.55, held-out-involved accuracy 0.3182, and
overall accuracy 0.4286. The visible-to-visible comparison already shows
that the strong construction result is sensitive to response recollection.
The narrow selection against observed candidate scores is a plausible
source of overfitting, although this single run does not isolate it from
response variability or grading sensitivity.

The verifier also uses clipped affine score calibration. Such calibration
can improve score scale without adding new task evidence. Because clipping
is applied at the individual-response level, it should not be assumed to
preserve aggregate model ordering in every possible response set.

\subsection{Claude-Opus-4.7: Revising Tasks, Repairing Judgments, and Rolling Back}
\label{app:case-claude}

The SWE-bench Claude-Opus-4.7 run illustrates several kinds of revision
within one session. Its initial 50-task benchmark uses unit tests,
bug-fix questions, reasoning problems, and rubric checks. The first
visible pairwise accuracy is 0.35. In the next round, the agent reports
that 42 tasks give every visible model full credit, and rewrites tasks
with stricter requirements and edge cases. Score spread increases, but
pairwise accuracy remains 0.35.

The subsequent changes go beyond difficulty alone. The agent replaces
several math and puzzle tasks with debugging problems, adds hidden
edge-case tests, and corrects erroneous reference answers. Iteration 4
reaches visible pairwise accuracy 0.55. The summaries describe errors
introduced by the benchmark itself, including expected outputs that were
inconsistent with JSON's conversion of tuples to lists. These examples
show that benchmark development includes debugging test specifications,
not only generating harder questions.

In iteration 6, the verifier gains a model-judged penalty for some
programmatically passing responses, and visible pairwise accuracy rises
to 0.65. Iteration 8 replaces a troublesome task with a permutation
validation problem and removes stale cached judgments associated with
the previous task version. Visible pairwise accuracy reaches 0.80 at
approximately 10.36 hours.

Later experiments fail to improve the ranking. The agent reports trying
new task replacements in iterations 9 and 10, observing regressions, and
restoring earlier artifacts. It eventually requests submission and ends
finalization at approximately 13.12 hours. The interval after iteration 8
therefore contains additional experiments and rollback decisions without
an increase in saved pairwise accuracy; it is not evidence that the agent
was inactive.

This case also exposes a verifier limitation. The final scoring code
retains a judgment cache indexed by task identifier and candidate name.
A cache lookup does not check whether the current response matches the
response originally judged, and candidates without an entry receive the
unmodified programmatic score. Consequently, fresh response collection
alone does not refresh every judgment embedded in arbitrary verifier
code. The final benchmark has overall confirmed pairwise accuracy 0.5952
and Spearman correlation 0.5881, but these metrics should not be read as
an audit of response-independent or model-independent grading.

\subsection{DeepSeek-V4-Pro: Targeted Revisions and Early Submission}
\label{app:case-pro-tb}

The Terminal-Bench DeepSeek-V4-Pro run starts with a broad, 114-task
snapshot and visible pairwise accuracy 0.3810. Its summary describes
near-ceiling performance on many conventional coding tasks. The agent
then changes the task family substantially, exploring self-reproducing
programs, polyglot programs, constrained programming, and other tasks
checked by executing the generated code.

The second snapshot contains only 17 selected tasks and is rejected for
insufficient count. Iteration 3 expands the submission to 41 tasks and
reaches valid visible pairwise accuracy 0.8095. In iteration 4, the agent
adds systems and bit-manipulation tasks, retains a no-vowel programming
task, fixes a scoring-branch error, and adds a timeout guard around
verification. The saved 49-task benchmark reaches pairwise accuracy
0.9048. These changes address both observed ranking errors and problems
that could prevent the verifier from finishing.

The next round tests whether additional systems-knowledge questions can
correct the remaining visible misorderings. The agent reports that they
do not help and leaves the task data and verifier unchanged from
iteration 4. It requests early submission rather than using the entire
remaining budget. Finalization yields 50 tasks, and the session finishes
after approximately 7.14 hours.

This is an example of targeted experimentation followed by a deliberate
stopping decision. It is not a demonstration that the selected task family
fully reproduces the reference evaluation. The final independent pass
obtains visible pairwise accuracy 0.6667, held-out-involved accuracy
0.5833, and overall accuracy 0.6222. The construction improvement and the
confirmation gap are both relevant to interpreting the run.

\subsection{DeepSeek-V4-Flash: Using the Remaining Budget Without a New Snapshot}
\label{app:case-flash-tb}

On Terminal-Bench, DeepSeek-V4-Flash completes its first iteration after
approximately 9.07 hours, with a 65-task snapshot and visible pairwise
accuracy 0.2857. The agent's summary describes several attempts to
increase difficulty, including more demanding specifications and
additional programming-language probes. It also reports fixing grading
bugs such as mutation of test inputs and incorrect reference values.

The second iteration consumes the remaining budget and is terminated at
approximately 16 hours. Its saved task prompts and verifier are unchanged
from the first snapshot, and visible pairwise accuracy remains 0.2857.
No dedicated finalization round is recorded. The final confirmed overall
pairwise accuracy is 0.4444.

The identical saved artifacts do not reveal how much exploratory work
occurred inside the second round. They establish the narrower observation
that almost seven additional hours did not produce an improved saved
benchmark. Alongside the early-submission example above, this suggests
that useful agent guidance should address the scope and duration of
experiments, keeping a complete submission available, and deciding when
further search is unlikely to help.

\section{Concrete Task and Scoring Examples}
\label{app:task-examples}

The following examples come from the final Terminal-Bench GPT-5.4 task
set. They illustrate the distinction between writing a reasonable prompt
and supplying a verifier that actually evaluates its requirements.

\subsection{A Task With an Explicit Scoring Branch}

The task \texttt{task\_001\_debug\_patch} asks for a correction to a
function that should return the index of the first repeated value:
\begin{quote}
\small
\begin{verbatim}
def first_repeat_index(items):
    seen = set()
    for i, value in enumerate(items):
        if value in seen:
            return value
        seen.add(value)
    return -1
\end{verbatim}
Explain the bug in one sentence, then provide a corrected implementation
in a single Python code block.
\end{quote}

The agent-authored reference answer requires identifying that the function
returns the repeated value instead of its index and providing a corrected
implementation. The verifier starts at 0.1, adds 0.45 for matching
sufficient explanatory keywords, and adds another 0.45 for a code block
containing a pattern matching \texttt{return i}. This branch responds to
answer content, but it is a heuristic rather than an execution-based
correctness check. Its presence should therefore be distinguished from
the strength of the evidence it uses.

\subsection{A New Task With a Reference Answer but No Scoring Branch}

The task \texttt{task\_007\_algorithm} asks:
\begin{quote}
\small
Write a Python function \texttt{merge\_intervals(intervals)} that merges
overlapping closed intervals such as
\texttt{[(1,3),(2,4),(7,8)]} into \texttt{[(1,4),(7,8)]}.
Include time complexity and handle empty input.
\end{quote}

Its reference answer requests a correct sorting-based merge algorithm,
valid Python code, and a complexity explanation. However, the submitted
verifier has no branch for this task identifier and does not otherwise
use its reference answer. Every nonempty answer receives 0.1. The
confirmation report records that value for all ten candidates.

Thus, the presence of a prompt and reference answer does not establish
that either affects evaluation. A task-level check could detect this
mismatch by inspecting dispatch coverage or comparing scores for answers
with clearly different content.

\subsection{The Same Failure Across a Different Task Type}

The task \texttt{task\_008\_code\_review} presents:
\begin{quote}
\small
\begin{verbatim}
def save_config(path, data):
    f = open(path, 'w')
    f.write(str(data))
    f.close()
\end{verbatim}
List the four most important issues, ordered by severity, then provide a
corrected version.
\end{quote}

Its reference answer mentions serialization, context management, encoding
and error handling, and possible atomicity concerns. This task also lies
outside the six verifier branches and receives the same default score.
The failure is therefore shared across implementation and review tasks;
adding more diverse prompts does not improve evaluation when those
prompts remain disconnected from the scorer.